\documentclass[letterpaper, 10 pt, journal, twoside]{IEEEtran}
\IEEEoverridecommandlockouts
\usepackage{cite}
\usepackage{amsmath,amssymb,amsfonts}
\usepackage{colortbl}
\usepackage[table, xcdraw, dvipsnames, svgnames, x11names]{xcolor}
\usepackage{amssymb}
\usepackage{multirow}
\usepackage{algorithmic}
\usepackage{graphicx}
\usepackage{textcomp}
\usepackage{xcolor}
\def\BibTeX{{\rm B\kern-.05em{\sc i\kern-.025em b}\kern-.08em
    T\kern-.1667em\lower.7ex\hbox{E}\kern-.125emX}}

\usepackage{algorithm}
\usepackage{array}
\usepackage[caption=false,font=normalsize,labelfont=sf,textfont=sf]{subfig}
\usepackage{stfloats}
\usepackage{url}
\usepackage{verbatim}

\usepackage{multirow}
\usepackage{hyperref}
\hypersetup{
    colorlinks=true,
    linkcolor=black,
    urlcolor=magenta,
    citecolor=black,
}

\newcommand{\PreserveBackslash}[1]{\let\temp=\\#1\let\\=\temp}
\newcolumntype{C}[1]{>{\PreserveBackslash\centering}p{#1}}

\begin{document}

\title{Detaching and Boosting: Dual Engine for Scale-Invariant Self-Supervised Monocular \\ Depth Estimation}

\author{Peizhe Jiang, Wei Yang, Xiaoqing Ye, Xiao Tan, and Meng Wu

\thanks{Peizhe Jiang and Meng Wu are with the school of Marine Science and Technology, Northwestern Polytechnical University, 
        Xi’an 710072, China 
        (e-mail: pz.jiang@mail.nwpu.edu.cn; wumeng@nwpu.edu.cn)}%
\thanks{Wei Yang, Xiaoqing Ye and Xiao Tan are with Department of Computer Vision Technology (VIS), Baidu Inc., 
        (e-mail: well\_young@163.com; yexiaoqing@baidu.com; tanxchong@gmail.com)
        }%
}

\maketitle

%%%%%%%%%%%%%%%%%%%%%%%%%%%%%%%%%%%%%%%%%%%%%%%%%%%%%%%%%%%%%%%%%%%%%%%%%%%%%%%%
\begin{abstract}
Monocular depth estimation (MDE) in the self-supervised scenario has emerged as a promising method as it refrains from the requirement of ground truth depth. Despite continuous efforts, MDE is still sensitive to scale changes especially when all the training samples are from one single camera. Meanwhile, it deteriorates further since camera movement results in heavy coupling between the predicted depth and the scale change. In this paper, we present a scale-invariant approach for self-supervised MDE, in which scale-sensitive features (SSFs) are detached away while scale-invariant features (SIFs) are boosted further. To be specific, a simple but effective data augmentation by imitating camera zooming process is proposed to detach SSFs, making the model robust to scale changes. Besides, a dynamic cross-attention module is designed to boost SIFs by fusing multi-scale cross-attention features adaptively.
Extensive experiments on the KITTI dataset demonstrate that the detaching and boosting strategies are mutually complementary in MDE and our approach achieves new State-of-The-Art performance against existing works from 0.097 to 0.090 w.r.t absolute relative error. The code is available at \url{https://github.com/AttackonMuggle/DaB_NET0}.
\end{abstract}


\IEEEpeerreviewmaketitle

\section{INTRODUCTION}
Monocular depth estimation (MDE) is a critical but challenging computer vision (CV) task, which has a wide range of applications in augmented reality and autonomous driving. With the surge of convolutional neural networks (CNN), most supervised approaches \cite{Fu2018,bhat2021adabins} have achieved leading performance. Nevertheless, ground truth (GT) of depth annotations is labor-intensive due to data sparsity and depth-sensing devices cost. 

\begin{figure}[ht]
    \centering
    \includegraphics[width=\linewidth]{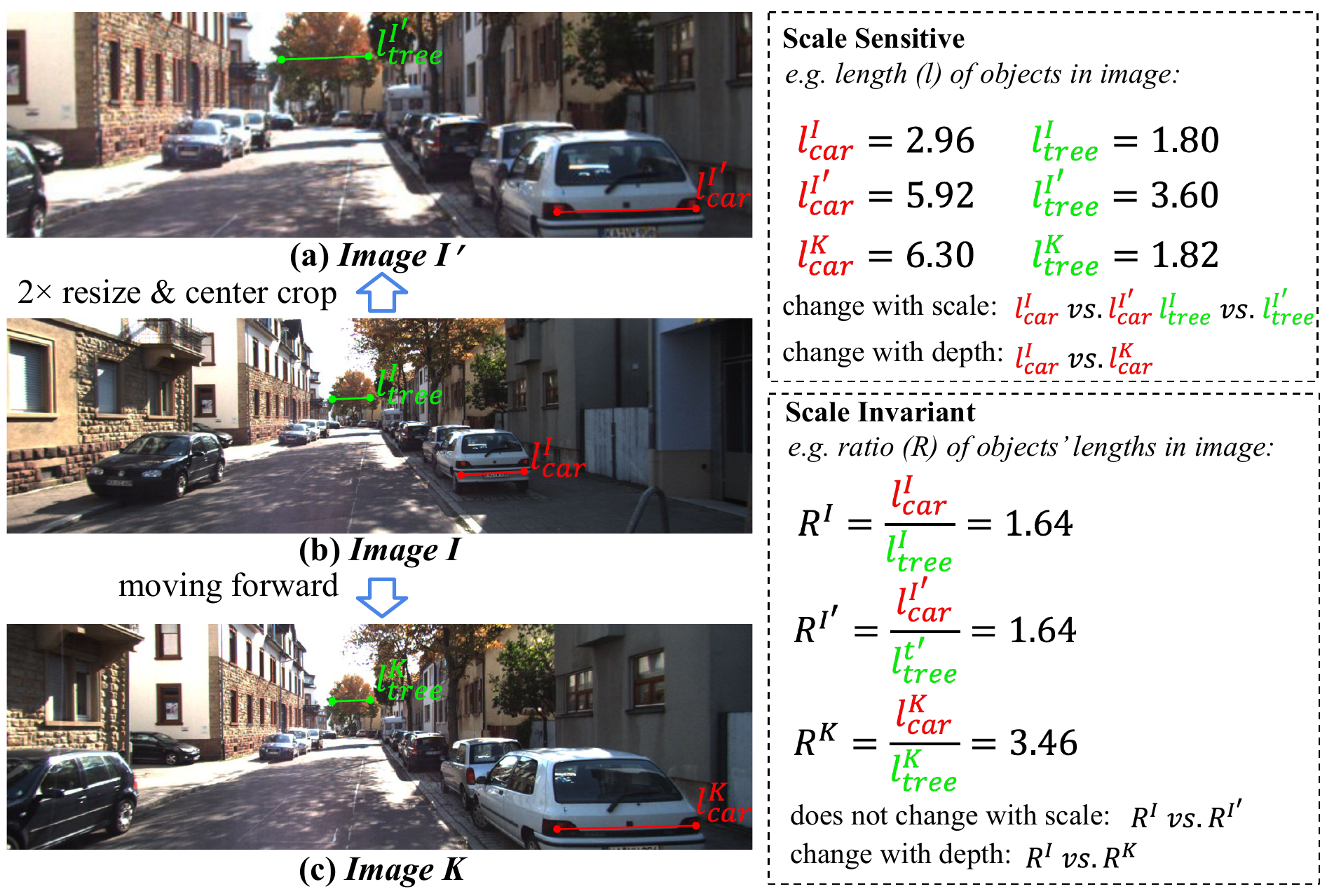}
    \caption{Illustration on the influence of scale changes to the features. Image \emph{$I'$} is an enlarged version of Image \emph{$I$} imitating the scene captured by the camera with a larger focal length from the same viewpoint, and hence the depths of the same object in both Image \emph{$I$} and Image \emph{$I'$} should be the same. While Image \emph{$K$} is captured after the camera moves forward from Image \emph{$I$}, the depths of the same object in Image \emph{$I$} and Image \emph{$K$} are different. 
    }
    \label{fig:ScaleAmbiguity}
\end{figure}

Recently, self-supervised approaches have gained more attention, because significant progress has been made on unsupervised learning of depth and ego-motion from unlabeled monocular video \cite{godard2019digging,shu2020feature}, stereo pairs \cite{gonzalez2021plade,chen2021revealing} or a combination of both. The task tends to predict the depth by exploiting the geometrical relations between target and source views in the training data. Despite the less requirement for data preparation, self-supervised approaches are still striving for a decent performance, especially when compared with their supervised counterparts. Most of previous works made large efforts to address limitations of the photometric difference by adding further auxiliary constraints \cite{shu2020feature,lee2021learning,gonzalezbello2020forget} since the supervision from view synthesis presents a dilemma, meaning strict pixel-wise correspondence between views.

Different from aforementioned works, we find that the change of depth is often accompanied by the scale change in camera movement, which is believed to be the very reason for severe performance degradation in MDE. By the term "scale", we mean the visual cues related with the appearance of an object in the image, which is determined jointly by the distance, camera focal length and the physical size of the object. 
To better illustrate this problem in Fig.~\ref{fig:ScaleAmbiguity}, we coarsely classify the visual cues for depth estimation into scale-sensitive features (SSFs) and scale-invariant features (SIFs). 
SSFs refer to the features which are easily influenced by both scale and depth in images, such as the pixel-wise length of the object in images. To put it vividly, the tree and car in Fig.~\ref{fig:ScaleAmbiguity} (a) appear larger in size and closer in distance than those in Fig.~\ref{fig:ScaleAmbiguity} (b) while they come from the same picture actually, which indicates the estimated depths in both cases should be equivalent in theory. On the contrary, other features which remain constant against the scale change but merely vary with depth are termed as scale-invariant features (SIFs), such as the ratio of pixel-wise lengths of two objects in the image.
In Fig.~\ref{fig:ScaleAmbiguity}, $R^I$ and $R^{{I}'}$ are equal as they are from the same image; while $R^K$ is different since the objects in image $I$ and image $K$ do have different depths. As is pointed out, this key observation motivates us to extract more SIFs but less SSFs for robust and reliable depth estimation. To fulfill the goal, we first resort to a novel method of camera zoom data augmentation (CZDA) to detach SSFs as much as possible, enabling the network to focus more on SIFs. After detaching SSFs, SIFs can be further enhanced via a dynamic cross-attention (DCA) module, which makes full use of multi-scale features by cross-attention and dynamic fusion. 

To summarize, our contributions are three-folds as follows:
\begin{itemize}
    \item The naive concepts of SSFs and SIFs are introduced for the first time in self-supervised monocular depth estimation. 
    \item A SIF-based MDE method consisting of detaching and boosting is proposed in a plug-and-play manner for various self-supervised frameworks. The experiments indicate the effectiveness and complementarity of these two measures.
    \item We achieve new SOTA performance on the KITTI benchmark, which outperforms all the previous self-supervised MDE methods by a large margin.
\end{itemize}

\label{sec:intro}

\section{RELATED WORK}

\subsection{Supervised Depth Estimation}
Methods falling into this category require ground truth annotations of depth, which are usually cumbersome. Recent years have witnessed the prosperity of CNN since the breakthrough work of \cite{eigen2014depth} achieves overwhelming performance over traditional methods.
DORN \cite{Fu2018} considers depth estimation as a classification task, in which depth discretization needs to be conducted at intervals for both GT and estimation. 
In another vein, Miangoleh et al. \cite{miangoleh2021boosting} proposed to boost existing MDE models by merging content-adaptive multi-scale estimations for high-resolution depth maps. To avoid over-fitting, several works \cite{Fu2018,eigen2014depth,laina2016deeper} utilized scale-related data augmentation for depth estimation. Despite the continuous and thriving progress made so far, supervised methods are inherently defamed in the thirst for a huge amount of densely annotated data, which is a formidable task compared with other CV tasks.

\subsection{Self-Supervised Depth Estimation}
As an alternative, self-supervised methods learn to predict depth without labels, in which the self-supervision can be easily obtained by clues like geometric constraints between multiple views. Garg et al. \cite{garg2016unsupervised} first applied a self-supervised training method with stereo pairs, in which photometric consistency loss using $L_2$ norm usually generates blurry depth maps. To solve this problem, Godard et al. \cite{godard2017unsupervised} replaced it with SSIM and $L_1$ norm to yield better results. From then on, a variety of successive works have developed further in terms of objective functions \cite{watson2019self,zhu2020edge} or model architectures \cite{gonzalez2021plade,chen2021revealing}.

Another stream of works predicts depth through camera ego-motion derived from monocular video, which is less demanding in data preparation but more challenging due to the unknown camera pose and moving objects.
Many attempts have been made, including minimum reprojection loss \cite{godard2019digging}
for dealing with occlusion, feature-metric loss \cite{shu2020feature} for enhancing textureless regions, and depth factorization and residual pose estimation for indoor environments \cite{ji2021monoindoor}. In addition, other methods try to exploit semantic information from pre-trained module \cite{jung2021fine} or semantic labels \cite{klingner2020self} to enforce depth consistent over dynamic regions or near the object boundaries. When it comes to the problem of scale ambiguity, several works attempt to enforce scale consistency by formulating scale-consistent geometric constraints between multiple views \cite{mahjourian2018unsupervised,wang2021can}. In this study, we aim to fully utilize scale-invariant features for depth estimation, based on a previously overlooked observation that severe coupling of the predicted depth with the scale change will result in degraded performance. Therefore, we seek a scale-invariant approach by detaching SSFs and boosting SIFs, which guides the network to predict depth robust to scale change.

\subsection{Self-attention}

The concept of attention started its dominance in natural language processing (NLP), and later in computer vision with its early success in CNN and later prosperity in Transformer. For depth estimation, Li et al. \cite{li2021revisiting} combined with transformer to replace the original network for stereo matching. Johnston et al. \cite{johnston2020self} explored non-contiguous image regions as a context for estimating similar depth. Yan et al. \cite{yan2021channel} made the DepthNet to better understand the 3D structure of the whole scene and Jung et al. \cite{jung2021fine} designed a cross-task attention module for refining depth features more semantically consistent. Most self-attention based methods mentioned above excel in capturing long-range dependencies and aggregating discriminative features, which guarantees a steady performance increase. Bearing the advantage of self-attention in mind, we further enhance SIFs hierarchically by a newly-designed dynamic cross-attention.

\section{METHOD}
\label{sec:method}

\begin{figure*}[ht]
    \centering 
	\includegraphics[width=\textwidth]{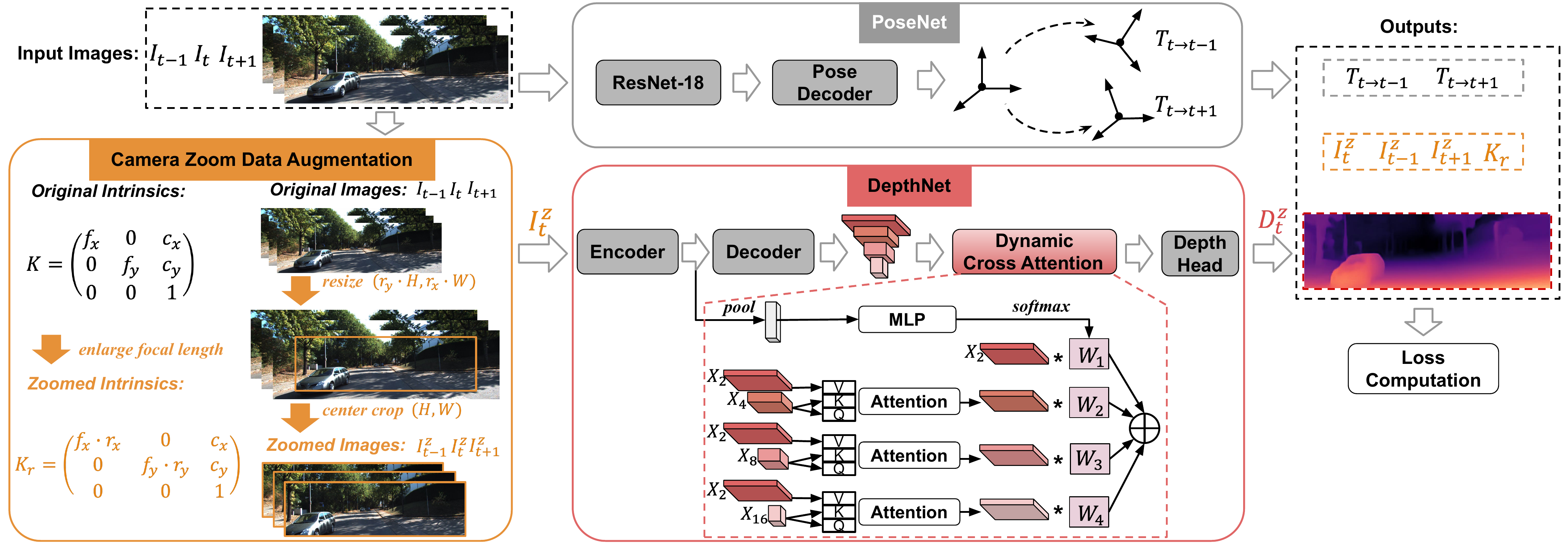}
	\caption{The overview of our pipeline. First the input camera intrinsic matrix $K$ and image sequence $(I_{t-1},I_{t},I_{t+1})$ are zoomed as $K_r$ and $(I_{t-1}^z,I_{t}^z,I_{t+1}^z)$ by camera zoom data augmentation. Then original image sequences are fed into the PoseNet to predict the relative poses $T_{t \rightarrow t-1}$ and $T_{t \rightarrow t+1}$. Taking the zoomed image $I_{t}^{z}$ as input, DepthNet outputs the depth map $D_t^z$. Zoomed data, relative poses and predicted depth map are used for training PoseNet and DepthNet.}
	\label{fig:NetworkArchitecture}
\end{figure*}

In this section, we first review the general framework for self-supervised monocular depth estimation, which consists of a DepthNet and a PoseNet. Then, we describe in detail how to integrate two core components of our SIF based approach, i.e., camera zoom data augmentation for detaching SSFs and dynamic cross-attention for boosting SIFs, into the whole framework seamlessly.

As is illustrated in Fig.~\ref{fig:NetworkArchitecture}, the image sequences are first transformed by camera zoom data augmentation before feeding to the DepthNet. It is assumed that SSFs can be eliminated to some degree by image zooming operation. Then a dynamic cross-attention module is involved to further encourage the network to predict depth from SIFs. Moreover, we leave PoseNet unchanged to get relative poses. With estimated depth and relative camera pose, a reprojected target image from a source image can be reconstructed by pixel-wise correspondence and thus act as the self-supervision signal.

%%%%%%%%%%%%%%%%%%%%%%%%%%%%%%%%%%%%%%%%%%%%%%%%%%%%%%%%%%%%%%%%%%%
\subsection{Self-supervised Monocular Depth Estimation}
\label{sec:previous}

Given consecutive RGB images $I_t$ and $I_s$, $s \in \{t-1, t+1\}$, and camera intrinsic matrix $K \in \mathbb{R}^{3 \times 3}$, we can derive the depth map $D_t$ of $I_t$ by DepthNet. Using monocular video as training, we need an auxiliary PoseNet to predict the relative pose $T_{t \rightarrow s}$ between $I_t$ and $I_s$. Then we can warp the image $I_s$ to $I_t$, following \cite{zhou2017unsupervised} by:

\begin{equation}
  I_{s \rightarrow t} = I_{s} \left \langle K T_{t \rightarrow s} D_t K^{-1}p_t \right \rangle ,
\label{eq:reproject}
\end{equation}
where $p_t$ represents homogeneous pixel coordinates of $I_t$, and reconstruction process \cite{zhou2017unsupervised} is mainly used in $\left \langle  \right \rangle $. $I_{s \rightarrow t}$ is the reprojected image from $I_s$ to $I_t$. 
Following previous studies \cite{godard2019digging,zhou2017unsupervised}, we stand on the foundation of structure from motion and construct the photometric loss $L_{ph}$, minimizing the discrepancy between $I_t$ and $I_{s \rightarrow t}$ to optimize DepthNet and PoseNet. In detail, $L_{ph}$ consists of the structural similarity index measure (SSIM) \cite{wang2004image} and $L_1$ loss:
\begin{equation}
\resizebox{0.438\textwidth}{!}{$
  L_{ph}(I_t, I_{s \rightarrow t}) = \alpha \frac{1-SSIM(I_t, I_{s \rightarrow t})}{2} + \beta \left | I_t - I_{s \rightarrow t} \right |,$}
\label{eq:photometric loss}
\end{equation}
where SSIM computes over a $3\times3$ pixel window, with hyper-parameters $\alpha$ and $\beta$. 
In addition, we apply minimum reprojection loss to deal with occlusions, and auto-mask to ignore static pixels where no relative camera motion is observed in monocular training \cite{godard2019digging}.

Following \cite{wang2018learning}, the edge-aware smoothness loss $L_{sm} $ is also added to prevent shrinking of the estimated depth:

\begin{equation}
L_{sm}^{i}=\sum_{p}e^{-\left | \nabla^{i} I\left ( p \right )   \right |_{1}  } \left | \nabla^{i} \hat{D_{t}} \left ( p \right )   \right |_{1},
\label{eq:smooth loss}
\end{equation}
where $\tilde{ D_{t}} = D_{t} / \bar{D_{t} } $ is the mean-normalized inverse depth. 

Finally, the loss function of our baseline is the combination of the reprojection loss and the smooth loss:

\begin{equation}
  L = \lambda L_{ph} + \mu \left ( L_{sm}^{1} + L_{sm}^{2} \right ),
\label{eq:total loss}
\end{equation}
where $\lambda $ and $\mu $ are used to balance their contributions. 

%%%%%%%%%%%%%%%%%%%%%%%%%%%%%%%%%%%%%%%%%%%%%%%%%%%%%%%%%%%%%%%%%%%%%%%%%%%%%%%%%%%%

\subsection{Camera Zoom Data Augmentation for Detaching SSFs}
\label{sec:augmentation}

As is demonstrated in Fig.~\ref{fig:ScaleAmbiguity}, features are coarsely classified into SSFs and SIFs by our definition. 
For SSFs, the appearance features of the same instance under different depth ranges can vary conspicuously in monocular image sequences, leading to the strong correlation between depth estimation and object scales in the image.
Inspired by the intuitive way humans perceive distant objects, we propose to mathematically adjust the focal length to decouple SSFs from the raw features and thus leave SIFs intact, which is the very essence of our camera zoom data augmentation. Specifically, the new data augmentation tries to imitate camera zooming where the focal length and resulting images are changed simultaneously such that the resulting depth is kept unchanged.

Suppose the original input of our pipeline is $I \in \mathbb{R}^{3\times H\times W}$, where $H$ and $W$ are the height and width of the original image. Camera intrinsic matrix is $K$ with focal length $f_x, f_y$ and principal point $c_x, c_y$. 
Here we modify $K$ to get the zoomed intrinsic matrix $K_r$ as follows:

\begin{equation}
   K_{r} = K\cdot r  = \begin{bmatrix}
 f_{x} \cdot  r_{x} & 0 & c_{x} \\
 0 & f_{y} \cdot  r_{y} & c_{y} \\
 0 & 0 & 1
\end{bmatrix}
,
\label{eq:zoom in}
\end{equation}
where $r_{x}$ and $r_{y}$ denote the zoom ratio along width and height. Note that the ratios we adopt for each image are randomly generated with a given probability at each iteration.
Correspondingly, we need to resize $I$ to get image ${I}' \in \mathbb{R}^{3\times h\times w}$, where $ h = H \cdot r_{y} $, $ w = W \cdot r_{x} $, and then center-crop ${I}'$ to the size of $(H,W)$ with the principal point as the center so that the final image we use is the zoomed image $I^{z} \in \mathbb{R}^{3\times H\times W}$.

Since the focal length has been changed, the images need to be resized and center-cropped correspondingly to ensure that the same 3D points in the world coordinate can correspond to the same pixels in $I$ and $I^z$.
To be specific, for a 3D point $S:[X,Y,Z]^T$ in camera coordinate system, $p:[u,v]^T$ is the pixel in image $I$ corresponds to $S$, $p_r:[u_r,v_r]^T$ is the pixel in image $I^z$ corresponds to $S$, so they have:
\begin{equation}
\resizebox{0.438\textwidth}{!}{$
    \begin{bmatrix}
    u_r-c_x \\
    v_r-c_y \\
    1
    \end{bmatrix}
    =
    \begin{bmatrix}
    f_x\cdot  r_x & &  \\
     & f_y\cdot  r_y &  \\
     & & 1
    \end{bmatrix} \cdot 
    \begin{bmatrix}
    X/Z \\
    Y/Z \\
    1
    \end{bmatrix}
    =
    \begin{bmatrix}
    (u-c_x)\cdot  r_x \\
    (v-c_y)\cdot  r_y \\
    1
    \end{bmatrix}$.}
    \label{eq:same_depth}
\end{equation}
Equation (\ref{eq:same_depth}) is exactly the resize and center-crop process from $p$ to $p_r$. It proves that our zoom process can keep the 3D geometric property unchanged, i.e., CZDA will not change the geometric relationship between objects and cameras.

The camera zoom process does not change the pose of adjacent frames so we use the original images as inputs of PoseNet, since the full image benefits to estimating pose by providing more context knowledge. Here (\ref{eq:reproject}) becomes:
\begin{equation}
   I_{s \rightarrow t}^{z} = I_{s}^{z} \left \langle K_{r} T_{t \rightarrow s} D_t^{z} K_{r}^{-1}p_{t}^{z} \right \rangle ,
\label{eq:reproject_zoom}
\end{equation}
where $I_{s}^{z}$ is the zoomed $I_{s}$, $D_{t}^{z}$ is the depth map of $I_{t}^{z}$, and $p_{t}^{z}$ is pixel coordinates of $I_{t}^{z}$. So the loss function in (\ref{eq:total loss}) is computed between $I_{t}^{z}$ and $I_{s \rightarrow t}^{z}$.

Based on the above deduction, it is assumed that the estimated depth is the unique factor influencing the reconstruction loss in (\ref{eq:photometric loss}) when feeding the data generated by CZDA. As a 3D geometry-driven method, the DepthNet is enforced to predict the same depth for different zoomed images so that the phtotometric loss in (2) can converge. In other words, SSFs can be detached from SIFs as much as possible during the training process, because the DepthNet is guided to predict the depth with more scale-insensitive features. 
Note that our CZDA is quite different from previous scale-related augmentation \cite{eigen2014depth,laina2016deeper} applied in the supervised methods, which changes the depth according to the zoom ratio. It is indicated that in self-supervised methods traditional data augmentation 
will break the strict pixel correspondence between views, thereby worsening the performance \cite{peng2021excavating}. After detaching SSFs in this way, SIFs play a dominant role for DepthNet, and the predicted depth is thereby believed to be more robust to scale change.

%%%%%%%%%%%%%%%%%%%%%%%%%%%%%%%%%%%%%%%%%%%%%%%%%%%%%%%%%%%%%%%%%%%%%%%%%%%%%%%%%%%%

\begin{figure}[ht]
    \centering
    \includegraphics[width=0.48\textwidth]{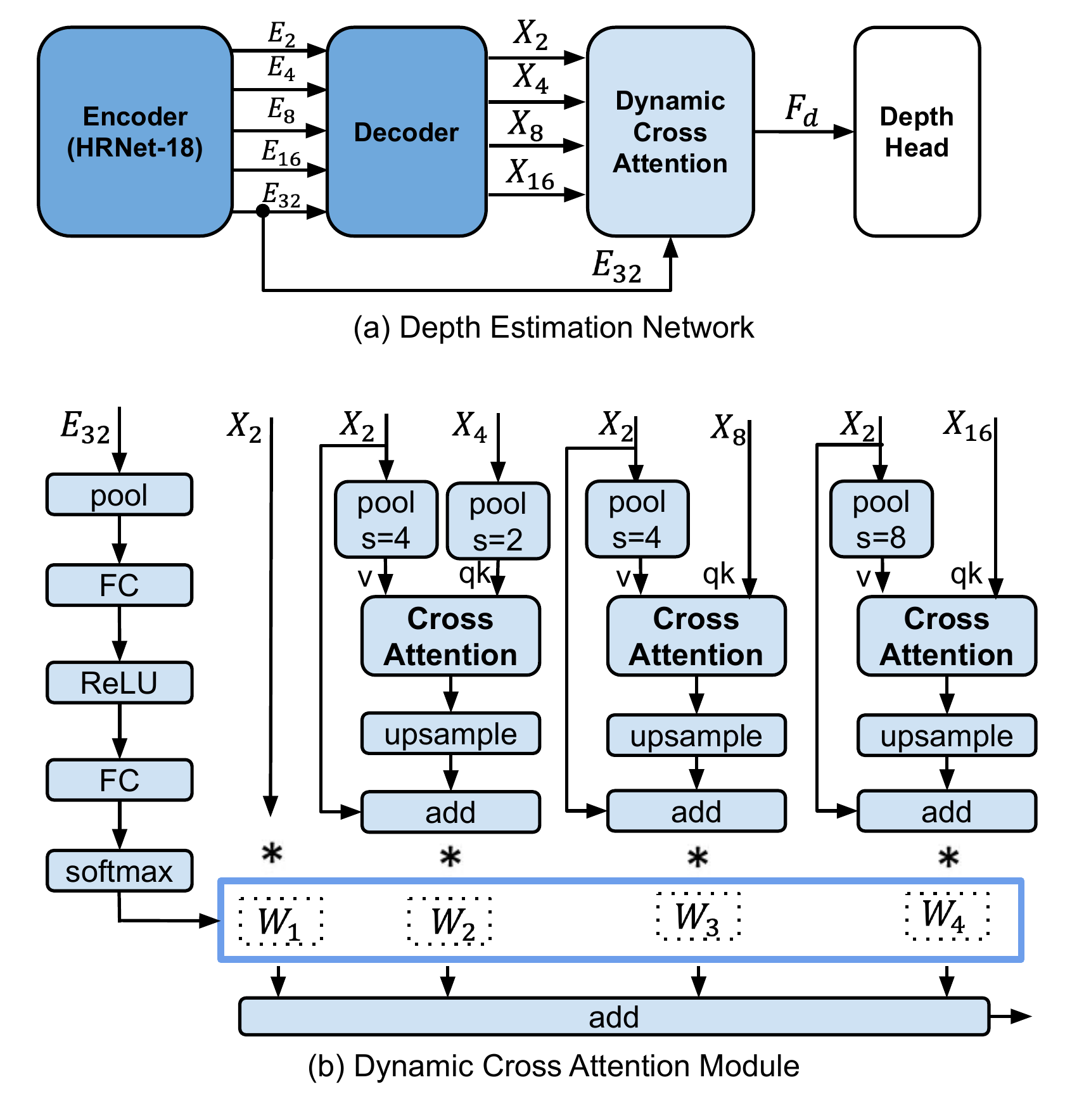}
    \caption{
    Detail of our dynamic cross-attention. The encoder and decoder structure is the same as DIFFNet.
    }
    \label{fig:SIFB detail}
\end{figure}

\subsection{Dynamic Cross-attention for Boosting SIFs}
\label{sec:block}

After detaching SSFs by CZDA, we expect to further boost SIFs, rendering the network more robust to scale change. 
As observed in previous works \cite{yan2021channel,sagar2022monocular}, both attention models and feature fusion strategies across various scales are beneficial to construct more reliable features for depth estimation. Therefore, we propose a dynamic cross-attention module which takes full advantage of multi-scale features by cross-attention and dynamic fusion.

Given the feature maps $X_{2}$, $X_{4}$, $X_{8}$ and $X_{16}$ from the decoder of DepthNet in Fig.~\ref{fig:SIFB detail} (a), $X_{i}$, $i\in [4,8,16]$ is supposed to be relatively global in a hierarchical way. Inspired by classic self-attention formula \cite{wang2018non}, we then define cross-attention as follows:
\begin{equation}
\resizebox{0.438\textwidth}{!}{$
    A\left ( X_i, X_2 \right ) = X_2+softmax\left ( Q\left ( X_i \right )^{T}K\left ( X_i \right ) \right ) V\left ( X_2 \right ),$}
\label{eq:attention}
\end{equation}
where the query $Q(\cdot)$, key $K(\cdot)$, and value $V(\cdot)$ represent $1\times1$ convolution preceded by pooling and followed by up-sampling to facilitate computation. The attention map $softmax\left ( Q\left ( X_i \right )^{T}K\left ( X_i \right ) \right )$ in Eq. (8) indicates the structural self-similarity of $X_i$. To obtain cross-attention features hierarchically, we feed the multi-scale features, like $X_{4}$, $X_{8}$ and $X_{16}$, to $Q$ and $K$ of cross-attention blocks correspondingly. For $V$, $X_{2}$ is kept constant. Pooling with adaptive stride beforehand is used to ensure the same feature dimension of two inputs for cross-attention block while up-sampling afterward is to match the dimension of $X_2$. As $i$ varies from 4 to 16, $X_{i}$ becomes more global and thus the cross-attention features $A\left ( X_i, X_2 \right )$ capture more coarse-grained structure, but less fine-grained details.

With these cross-attention features, our next concern is how to fuse them for enhanced SIFs. Here, we follow \cite{chen2020dynamic} to generate dynamic weights for feature fusion indicated by:

\begin{equation}
   W=softmax(MLP(GAP(E_{32}))),
\label{eq:dynamic weight}
\end{equation}
where $W \in \mathbb{R}^{1\times 4}$, $E_{32}$ is the highest level feature of the encoder, $MLP$ is multi-layer perception, and $GAP$ stands for global average pooling. The final enhanced SIFs of $F_d$ are then defined as:
\begin{table*}
   \caption{Comparison results on the KITTI dataset. Best results are in {\bf bold}, second best are \underline{underlined}.
   For {\color[HTML]{E9967A}{\bf red}} metrics, lower is better, and higer is better for {\color[HTML]{B9CEFA}{\bf blue}} metrics. 
   Abbreviation in Data column: D refers to supervised methods with GT depth supervision, $\mathrm{D}^\dagger$ uses auxiliary depth supervision from SLAM, $\mathrm{D}^\ast$ uses auxiliary depth supervision from synthetic depth labels, +Sem means additional supervision from semantic labels or pre-trained segmentation network, S refers to training on stereo images and M for training by monocular videos.
%   and MS for training by binocular videos. 
   PP represents post-processing \cite{godard2017unsupervised} for most methods. 
   }
   \begin{center}
   \resizebox{0.99\linewidth}{!}{
   \begin{tabular}{|l|c|c|c||c|c|c|c|c|c|c|c|}
   \hline
   Method & PP & Data & $H \times W$ & \cellcolor[HTML]{E9967A}Abs Rel & \cellcolor[HTML]{E9967A}Sq Rel & \cellcolor[HTML]{E9967A}RMSE & \cellcolor[HTML]{E9967A}RMSE log & \cellcolor[HTML]{B9CEFA}$\delta < 1.25$ & \cellcolor[HTML]{B9CEFA}$\delta < 1.25^2$ & \cellcolor[HTML]{B9CEFA}$\delta < 1.25^3$ \\
%   training by D
  \hline 
  Eigen \cite{eigen2014depth} &            & D & $184 \times 612$ & 0.203 & 1.548 & 6.307 & 0.282 & 0.702 & 0.890 & 0.890 \\
  Yang \cite{Yang2018Deep}  & \checkmark & $\mathrm{D}^\dagger$S & $256 \times 512$ & 0.097 & 0.734 & 4.442 & 0.187 & 0.888 & 0.958 & 0.980 \\
  Luo \cite{luo2018single} &            & $\mathrm{D}^\ast$DS & $192 \times 640$ crop & 0.094 & 0.626 & 4.252 & 0.177 & 0.891 & 0.965 & 0.984 \\
  DORN \cite{Fu2018} RestNet  &            & D & $385 \times 513$ crop & \underline{0.072} & \underline{0.307} & \underline{2.727} & \underline{0.120} & \underline{0.932} & \underline{0.984} & \underline{0.994} \\
  AdaBins \cite{bhat2021adabins}  &       & D & $352 \times 704$ & \bf{0.058} & \bf{0.190} & \bf{2.360} & \bf{0.088} & \bf{0.964} & \bf{0.995} & \bf{0.999} \\
  \hline
% %   training by S
  \hline
  Monodepth \cite{godard2017unsupervised} & \checkmark & S & $256 \times 512$ & 0.138 & 1.186 & 5.650 & 0.234 & 0.813 & 0.930 & 0.969 \\
  Monodepth2 \cite{godard2019digging} & \checkmark & S & $320 \times 1024$ & 0.105 & 0.822 & 4.692 & 0.199 & 0.876 & 0.954 & 0.977 \\
  DepthHints \cite{watson2019self} & \checkmark & S & $320 \times 1024$ & 0.096 & 0.710 & 4.393 & 0.185 & 0.890 & 0.962 & 0.981 \\
%   FALnet \cite{gonzalezbello2020forget} &  & S & $192 \times 640$ crop & 0.097 & 0.590 & 3.991 & 0.177 & 0.893 & 0.966 & 0.984 \\
  SingleNet \cite{chen2021revealing} & \checkmark & S & $320 \times 1024$ & 0.094 & 0.681 & 4.392 & 0.185 & 0.892 & 0.962 & 0.981 \\
  FALnet \cite{gonzalezbello2020forget} & \checkmark & S & $192 \times 640$ crop & 0.094 & \underline{0.597} & \bf{4.005} & \underline{0.173} & \underline{0.900} & \bf{0.967} & \bf{0.985} \\
  EPCNet \cite{peng2021excavating} & \checkmark & S & $320 \times 1024$ & \underline{0.091} & 0.646 & 4.207 & 0.176 & \bf{0.901} & \underline{0.966} & 0.983 \\
  EdgeDepth \cite{zhu2020edge} & \checkmark & S+Sem & $320 \times 1024$ & \underline{0.091} & 0.646 & 4.244 & 0.177 & 0.898 & \underline{0.966} & 0.983 \\
  PLADE-Net \cite{gonzalez2021plade} &  & S & $192 \times 640$ crop & 0.092 & 0.626 & 4.046 & 0.175 & 0.896 & 0.965 & \underline{0.984} \\
  PLADE-Net \cite{gonzalez2021plade} & \checkmark & S & $192 \times 640$ crop & \bf{0.089} & \bf{0.590} & \underline{4.008} & \bf{0.172} & \underline{0.900} & \bf{0.967} & \bf{0.985} \\
%   training by M
  \hline
  Monodepth2 \cite{godard2019digging} & \checkmark & M & $320 \times 1024$ & 0.112 & 0.838 & 4.607 & 0.187 & 0.883 & 0.962 & 0.982 \\
  Johnston. \cite{johnston2020self} &  & M & $192 \times 640$ & 0.106 & 0.861 & 4.699 & 0.185 & 0.889 & 0.962 & 0.982 \\
  HR-Depth \cite{lyu2021hr} &  & M & $384 \times 1280$ & 0.104 & 0.727 & 4.410 & 0.179 & 0.894 & 0.966 & \underline{0.984} \\
  Featdepth \cite{shu2020feature} &  & M & $320 \times 1024$ & 0.104 & 0.729 & 4.481 & 0.179 & 0.893 & 0.965 & \underline{0.984} \\
  CADepth-Net \cite{yan2021channel} &  & M & $320 \times 1024$ & 0.102 & 0.734 & 4.407 & 0.178 & 0.898 & 0.966 & \underline{0.984} \\
  FSRE-Depth \cite{jung2021fine} & \checkmark & M+Sem & $192 \times 640$ & 0.102 & 0.675 & 4.393 & 0.178 & 0.893 & 0.966 & \underline{0.984} \\
  PackNet-SfM \cite{guizilini2020semantically} & \checkmark & M+Sem & $320 \times 1024$ & 0.100 & 0.761 & 4.270 & 0.175 & 0.902 & 0.965 & 0.982 \\
  DIFFNet \cite{zhou2021self} &  & M & $320 \times 1024$ & 0.097 & 0.722 & 4.345 & 0.174 & 0.907 & 0.967 & \underline{0.984} \\
  \bf{Ours} &  & M & $320 \times 1024$ & \underline{0.092} & \underline{0.640} & \underline{4.208} & \underline{0.169} & \underline{0.909} & \underline{0.969} & \bf{0.985} \\
  \bf{Ours} & \checkmark & M & $320 \times 1024$ & \bf{0.090} & \bf{0.597} & \bf{4.087} & \bf{0.167} & \bf{0.912} & \bf{0.970} & \bf{0.985} \\
  %   training by MS
  \hline
  \end{tabular}
   }
   \end{center}
   \label{tb:compare}
\end{table*}

\begin{equation}
\begin{split}
   F_{d} &= W_{1}X_{2}+W_{2}A(X_4,X_2)+W_{3}A(X_8,X_2)\\
   &+W_{4}A(X_{16},X_2).
\label{eq:dynamic output}
\end{split}
\end{equation}
With $F_{d}$, a depth head is added to predict the depth. We expect dynamic fusion to focus more on cross-attention features of coarse-grained structure for images in larger context since their layout is more important to depth estimation. Meanwhile, dynamic fusion is supposed to excite more fine-grained structural features in smaller context instead.

In fact, the statistical distribution of dynamic weights reveals that the weights $W_1$ and $W_2$ tend to be larger for a zoomed image, while $W_3$ and $W_4$ turn bigger for an original image, verifying the mechanism of DCA exactly. Due to the page limit, we will not provide the results in the paper. To elaborate it further, cross-attention generates multi-scale representation of the input features with hierarchical structure clue, and then the weights of the multi-scale features are dynamically adjusted for scale invariant features (SIFs) representation. 

As two measures of CZDA and DCS are orthogonal in nature, they are supposed to be complementary. The experiments prove that enhancing SIFs in this way is necessary for MDE after SSFs are detached.

\section{EXPERIMENTS}

In this section, we first make a comprehensive comparison with tens of SOTA methods. Then, we analyze the role of detaching SSFs, i.e., the significance of SIF's introduction. Finally, we perform extensive ablation studies on each component of our approach and validate the effectiveness and generalization of camera zoom data augmentation in other frameworks. 

\subsection{Dataset and Performance Metrics}
Our model was trained on the KITTI 2015 dataset \cite{geiger2012we}, which contains videos in 200 street scenes captured by RGB cameras, with sparse depth ground truths captured by Velodyne laser scanner. For training, we remove static frames by a pre-processing step suggested by \cite{zhou2017unsupervised}. We adopt the Eigen split of \cite{eigen2014depth} to divide KITTI raw data, resulting in 39,810 monocular triplets for training, 4,424 for validation and 697 for testing.

The performance is assessed by standard metrics, like absolute relative difference, square related difference, RMSE and log RMSE. For the methods trained on monocular videos, the depth is defined up to scale factor $\hat{s} $ during evaluation \cite{zhou2017unsupervised}, which is computed by:
\begin{equation}
   \hat{s}  = median\left (  D_{gt}  \right )/median\left (  D_{pred}  \right ) .
\label{eq:ratio}
\end{equation}
where the predicted depth maps $D_{pred}$ are multiplied by the computed scale factor $\hat{s}$ to match the median of the ground truth $D_{gt}$. The scale factor $\hat{s}$ here is also an important inspector of SSFs, and we will discuss it later.

\begin{figure*}[ht]
	\centering 
	\includegraphics[width=0.99\textwidth]{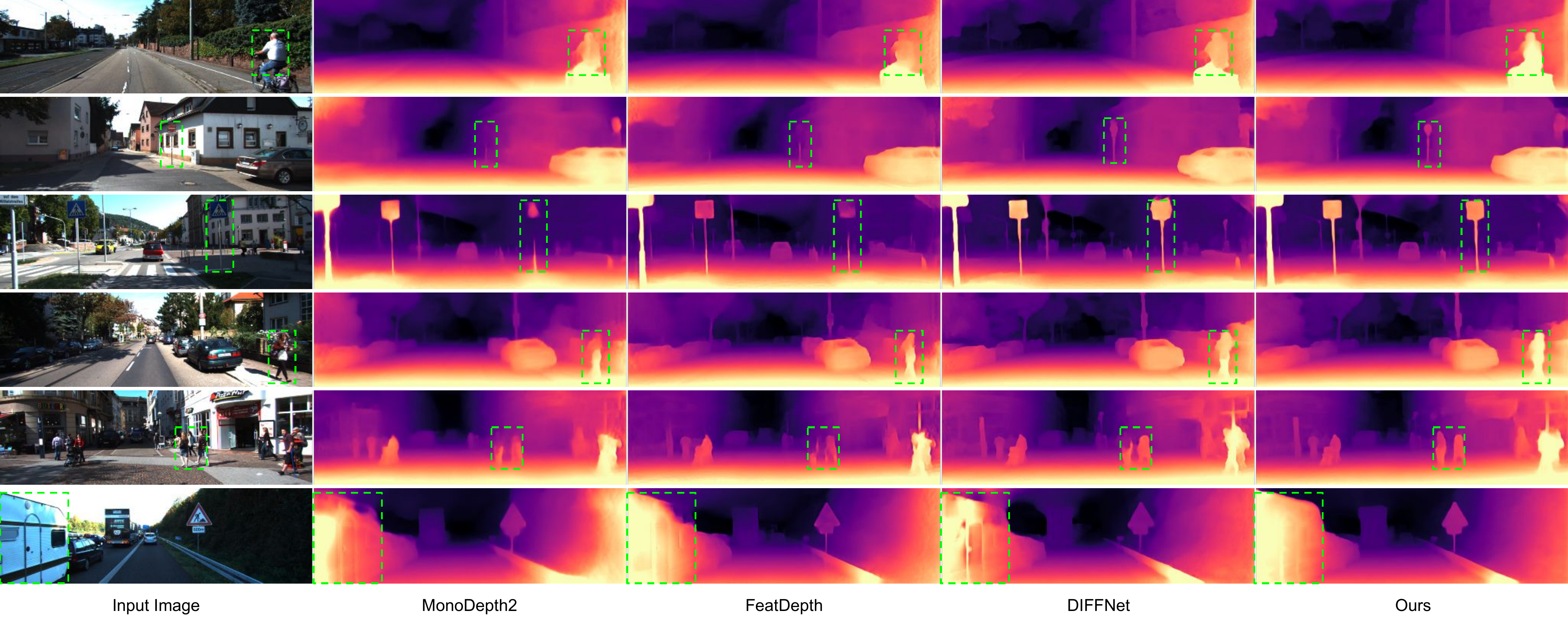}
	\caption{Qualitative visualization results on KITTI Eigen Test Split without post-processing. Our approach performs better on thinner objects such as signs and bollards, as well as being better at delineating difficult object boundaries like poles and humans.}
	\label{fig:Qualitative}
\end{figure*}

\subsection{Implementation Details}

Our DepthNet has a similar network architecture as DIFFNet \cite{zhou2021self}, which adopts HR-net \cite{wang2020deep} as the encoder and the channel-wise attention in the decoder. Unlike DIFFNet that uses multi-scale depth maps at training, we predict a single depth map with the same scale of input for loss computation in (\ref{eq:total loss}). In terms of PoseNet, we simply follow the architecture in \cite{shu2020feature}, considering the relative camera poses are fairly simple in outdoor environments (e.g., KITTI) 
%compared to indoor scenes 
\cite{ji2021monoindoor}. DepthNet and PoseNet respectively take the image sizes of $320 \times 1024$ and $192 \times 640$ as inputs.

We follow the setting in \cite{godard2019digging} for data pre-processing. The models are trained and tested on 4 Tesla V100. Adam optimizer with the default betas 0.9 and 0.999 is used. The learning rate starts from $1e^{-4}$, and then decays to half of the original at the 20th and 30th epoch \cite{shu2020feature}.

Unlike prior post-processing (PP) \cite{gonzalez2021plade} \cite{godard2017unsupervised}, we here adopt a novel zoom average post-processing, which fuses different depth estimation results from different zoom scales. We run the depth estimator three times to get three different depth maps, one for the original image and the other two for 1.33$\times$ and 2$\times$ zoomed images respectively. To be specific, the images are first resized to the sizes of $(1.33H, 1.33W)$ and $(2H, 2W)$, and then center-cropped to $(H, W)$ to get 1.33$\times$ and 2$\times$ zoomed images. Finally, we restore these depth maps to the original sizes by bilinear interpolation and average them directly in the corresponding region for the final output.

\subsection{Evaluation on KITTI}

Tab.~\ref{tb:compare} lists the quantitative results of various methods on the KITTI dataset \cite{eigen2015predicting}. During our evaluation, we cap depth to 80m per standard practice \cite{godard2017unsupervised}. Note that FAL-Net and PLADE-Net adopt multi-scale post-processing (PP) strategy \cite{gonzalez2021plade}, while our approach utilizes the proposed zoom average PP measure for better results. The table is split into several parts according to the supervision level, the post-processing, and the size of images. The results in the bottom block of Tab.~\ref{tb:compare} 
show that two variants of our approach significantly outperform all existing SOTA self-supervised methods using monocular video w.r.t. all the metrics.
% and higher results can be achieved with our zoom average PP. 
Higher results can be achieved with our zoom average PP, which verifies the superiority of our approach in aggregating the depths of images with different scales.
We also outperform recent methods with semantic label assisted \cite{jung2021fine} \cite{guizilini2020semantically}. Furthermore, our approach yields comparable or better results than all of self-supervised methods using stereo images. 
As is known, post-processing strategies generally lead to the growth of running time by fusing different depth results. Our zoom average PP will increase about $3\times$ the runtime, as it fuses three different depth estimation results.

\begin{table}[ht]
    \caption{Median scale factor change with and without camera zoom data augmentation (CZDA) on KITTI test set.}
    \centering
    \resizebox{1.0\linewidth}{!}{
    \begin{tabular}{|l||c|c|c|c|}
    \hline
       Scale Factor & Ours & Ours+CZDA & Monodepth2 & Monodepth2+CZDA \\
       \hline
        $median(\hat{s}_c)$ & 28.886 & 29.863 & 34.152 & 31.447 \\
        \hline
        $median(\hat{s}_z)$ & 52.536 & 30.424 & 63.672 & 31.126 \\
        \hline
    \end{tabular}
    }
    \label{tb:ratio}
\end{table}

In addition, Fig.~\ref{fig:Qualitative} illustrates the qualitative performance of our method against Monodepth2 \cite{godard2019digging}, Featdeapth \cite{shu2020feature} and DIFFNet \cite{zhou2021self} without post-processing. It is obvious that our approach achieves better performance on thinner objects and low-texture regions, like signs, bollards, and the side of bus, while retaining finer details like silhouettes of humans and poles. Our approach presents overall more structurally consistent and accurate depth over the entire scene.  

\begin{figure*}[ht]
	\centering 
	\includegraphics[width=0.99\textwidth]{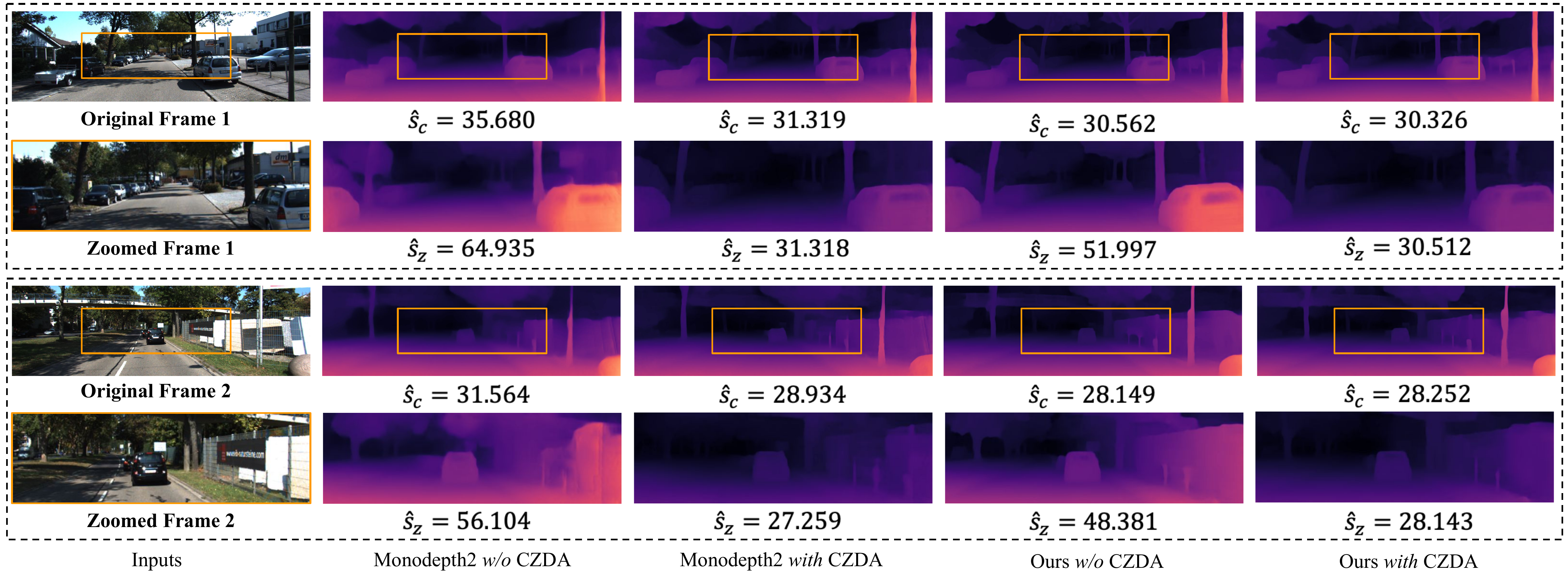}
	\caption{Visualization of scale factor change. CZDA is the abbreviation of camera zoom data augmentation. For generating appropriate visualization results, minimum depth and maximum depth are fixed as 0.1 and 100 respectively for all the depth maps. Zoomed frames correspond to the orange center crop box region in original frames. Scale factor are computed as in (\ref{eq:ratio}), $\hat{s}_c$ is the scale factor corresponding to the orange center crop box, and $\hat{s}_z$ is the scale factor of zoomed image. The GT used for scale factor computation are the GT of center crop regions.}
	\label{fig:Ratio}
\end{figure*}

\subsection{Analysis on Detaching Scale-Sensitive Features}
\label{sec:analysis}
% by Yang Wei
To validate the proposed detaching strategy, we resort to scale factor mentioned above. Self-supervised MDE merely gets relative depth, while scale factor $\hat{s}$ is used to adjust the predicted depth map to GT, indicating whether the method is insensitive to scale change. In other words, for both original image and zoomed image, scale factors of methods with camera zoom data augmentation are supposed to be constant.

We test Monodepth2 \cite{godard2019digging} and our method with/without CZDA for a qualitative comparison, as shown in Fig.~\ref{fig:Ratio}. $\hat{s}_c$ corresponds to the scale factor of the center crop region, and $\hat{s}_z$ is that of zoomed image.

\begin{table*}[h]
   \caption{Ablation results amongst variants of our model on the KITTI dataset. CZDA refers to camera zoom data augmentation and DCA refers to dynamic cross-attention module.}
   \centering
   \resizebox{0.75\textwidth}{!}{
   \begin{tabular}{|l|c||c|c|c|c|c|c|c|c|}
   \hline
   CZDA & DCA  & \cellcolor[HTML]{E9967A}Abs Rel & \cellcolor[HTML]{E9967A}Sq Rel & \cellcolor[HTML]{E9967A}RMSE & \cellcolor[HTML]{E9967A}RMSE log & \cellcolor[HTML]{B9CEFA}$\delta < 1.25$ & \cellcolor[HTML]{B9CEFA}$\delta < 1.25^2$ & \cellcolor[HTML]{B9CEFA}$\delta < 1.25^3$ \\
  \hline
   & & 0.096 & 0.676 & 4.280 & 0.173 & 0.905 & 0.968 & 0.984 \\
  \noalign{{\color{LightGray}\hrule height 0.4pt}}
%   & & 0.153 & 0.814 & 4.534 & 0.209 & 0.813 & 0.966 & 0.985 \\
%   \noalign{{\color{LightGray}\hrule height 0.4pt}}
   \checkmark & & 0.095 & 0.657 & 4.198 & 0.170 & 0.907 & 0.969 & 0.985 \\
  \noalign{{\color{LightGray}\hrule height 0.4pt}}
%   \checkmark & & 0.093 & 0.609 & 4.053 & 0.167 & 0.911 & 0.970 & 0.985 \\
%   \noalign{{\color{LightGray}\hrule height 0.4pt}}
   & \checkmark & 0.097 & 0.670 & 4.268 & 0.172 & 0.904 & 0.968 & 0.984 \\
  \noalign{{\color{LightGray}\hrule height 0.4pt}}
%   & \checkmark & 0.149 & 0.793 & 4.496 & 0.205 & 0.825 & 0.966 & 0.985 \\
%   \noalign{{\color{LightGray}\hrule height 0.4pt}}
   \checkmark & \checkmark & 0.092 & 0.640 & 4.208 & 0.169 & 0.909 & 0.969 & 0.985 \\
%   \noalign{{\color{LightGray}\hrule height 0.4pt}}
%   \checkmark & \checkmark & 0.090 & 0.597 & 4.087 & 0.167 & 0.912 & 0.970 & 0.985 \\
  \hline 
  \end{tabular}
  }
   \label{tb:ablation}
\end{table*}

\begin{table}[h]
   \caption{Results of our model with different zoom ranges.}
   \begin{center}
   \resizebox{0.97\linewidth}{!}{
   \begin{tabular}{|l||c|c|c|c|}
   \hline
    Zoom Range & \cellcolor[HTML]{E9967A}Abs Rel & \cellcolor[HTML]{E9967A}Sq Rel & \cellcolor[HTML]{E9967A}RMSE & \cellcolor[HTML]{B9CEFA}$\delta < 1.25$ \\
   \hline
   1$\sim $1.33 & 0.095 & 0.664 & 4.274 & 0.907 \\
   \noalign{{\color{LightGray}\hrule height 0.4pt}}
   1$\sim $2 & 0.092 & 0.640 & 4.208 & 0.909 \\
   \noalign{{\color{LightGray}\hrule height 0.4pt}}
   1$\sim $3.33 & 0.094 & 0.650 & 4.219 & 0.908 \\
   \noalign{{\color{LightGray}\hrule height 0.4pt}}
   1.25$\sim $3.33 & 0.094 & 0.663 & 4.228 & 0.907 \\
   \hline 
   \end{tabular}
   }
   \end{center}
   \label{tb:zoom_scale}
\end{table}

It is obvious that $\hat{s}_c$ has little difference with $\hat{s}_z$ for these two methods trained with CZDA, while the difference between $\hat{s}_c$ and $\hat{s}_z$ varies largely without CZDA. A similar conclusion can be drawn from the colorized depth maps, i.e., the center crop region color does not change when trained with camera zoom data augmentation. Furthermore, we perform this experiment on all the test set of KITTI with the results shown in Tab.~\ref{tb:ratio}. Experimental results verify that the proposed camera zoom data augmentation enables the network to detach SSFs, making the model robust to scale changes, and thus with our designed PP strategy, we can fuse different scales depths directly.

\subsection{Ablation Study}
In Tab.~\ref{tb:ablation}, we present a comprehensive ablation study of the key components of the proposed approach on KITTI. The addition of camera zoom data augmentation brings consistent performance increase w.r.t all metrics. Applying dynamic cross-attention module further improves the performance. It is interesting that directly applying DCA to the baseline yields no positive effect. The reason behind is that DCA plays the role of deeply excavating the initial features. Without CZDA for detaching SSFs, the network with DCA is still prone to predict depths from SSFs. 
On the other hand, the combination of CZDA and DCA gains a large performance improvement, which proves that the two components are complementary.

\begin{table}[h]
   \caption{Results of our camera zoom data augmentation on other self-supervised MDE methods on the KITTI dataset. CZDA refers to camera zoom data augmentation.}
   \resizebox{0.47\textwidth}{!}{
   \begin{tabular}{|l||c|c|c|c|}
   \hline
   Method & \cellcolor[HTML]{E9967A}Abs Rel & \cellcolor[HTML]{E9967A}Sq Rel & \cellcolor[HTML]{E9967A}RMSE &  \cellcolor[HTML]{B9CEFA}$\delta < 1.25$ \\
  \hline
  Monodepth2 & 0.115 & 0.882 & 4.701 & 0.879 \\
  \noalign{{\color{LightGray}\hrule height 0.4pt}}
  Monodepth2+CZDA& 0.106 & 0.789 & 4.537 & 0.891 \\
  \noalign{{\color{LightGray}\hrule height 0.4pt}}
  Featdepth & 0.104 & 0.729 & 4.481 & 0.893 \\
  \noalign{{\color{LightGray}\hrule height 0.4pt}}
  Featdepth+CZDA & 0.098 & 0.662 & 4.331 & 0.901 \\
  \hline
  \end{tabular}
  }
   \label{tb:effective}
\end{table}

\begin{table}[h]
    \caption{Comparison results of various attention modules on the KITTI dataset. CA refers to channel-wise attention, SA refers to spatial-attention, and Cross-A refers cross-attention.}
   \begin{center}
   \resizebox{0.97\linewidth}{!}{
   \begin{tabular}{|l||c|c|c|c|}
   \hline
   Method  & \cellcolor[HTML]{E9967A}Abs Rel & \cellcolor[HTML]{E9967A}Sq Rel & \cellcolor[HTML]{E9967A}RMSE & \cellcolor[HTML]{B9CEFA}$\delta < 1.25$ \\
  \hline
   CA \cite{yan2021channel} & 0.096 & 0.773 & 4.461 & 0.906 \\
  \noalign{{\color{LightGray}\hrule height 0.4pt}}
   SA \cite{zhou2021self} & 0.095 & 0.687 & 4.265 & 0.907 \\
  \noalign{{\color{LightGray}\hrule height 0.4pt}}
   Cross-A & 0.095 & 0.669 & 4.236 & 0.907 \\
  \noalign{{\color{LightGray}\hrule height 0.4pt}}
   DCA (ours) & 0.092 & 0.640 & 4.208 & 0.909 \\
  \hline 
  \end{tabular}
  }
  \end{center}
   \label{tb:ablation_att}
\end{table}

To investigate CZDA further, we examine the key factor of focal length. 
Here we set our camera zoom data augmentation with different zoom ranges i.e., the random range of $r_{x}$ and $r_{y}$ in (\ref{eq:zoom in}) and then validate our model with different settings. The results in Tab.~\ref{tb:zoom_scale} show that the zoom range from 1 to 2 works best. One possible reason is that 
a wider range will bring more quantization error while a narrower range undermines the potential of CZDA.

We also apply our camera zoom data augmentation in other frameworks \cite{godard2019digging,shu2020feature}, as is listed in Tab.~\ref{tb:effective}. The results demonstrate the effectiveness and generalization of the proposed CZDA, which can be utilized in a plug-and-play way for self-supervised MDE and bring considerable improvement. Considering the training efficiency, the loss function is computed on only one scale for Featdepth \cite{shu2020feature}.

Besides, we replace our DCA with other attention modules and give the comparison results on KITTT in Tab.~\ref{tb:ablation_att}. It's obvious that our DCA achieves the best performance, which is attributed to the dynamic fusion of cross-attention features.

\section{CONCLUSION}
In this paper, we bring forward a novel scale-invariant feature (SIF) based monocular depth estimation with a view to the observation that scale change matters. The proposed method consists of two measures of detaching and boosting. When it refers to detaching, the method invents a new data augmentation of camera zooming to detach SSF. Meanwhile, SIF is further boosted via a dynamic cross-attention module. Thorough experiments validate SIF and also indicate its striking superiority. In the future, we will explore the consistency constraint between depth maps from original image and zoomed image to improve the performance.

\bibliography{ieee_trans}
\bibliographystyle{IEEEtran}

\end{document}